\documentclass[journal]{IEEEtai}

\usepackage[colorlinks,urlcolor=blue,linkcolor=blue,citecolor=blue]{hyperref}

\usepackage{color,array}
\usepackage{color}
\usepackage{amsmath}
\usepackage{amssymb}
\usepackage{statex}
\usepackage{makecell}
\usepackage{booktabs}
\usepackage{multirow}
\usepackage{cite}
\usepackage{doi}
\usepackage{setspace}
\usepackage{tabularx}
\usepackage{color}
\usepackage{array}
\usepackage{graphicx}

\begin{document}

\title{Stereo Image Coding for Machines with Joint Visual Feature Compression} 

\author{Dengchao Jin,
	Jianjun~Lei,~\IEEEmembership{Senior~Member,~IEEE,}
	Bo Peng,~\IEEEmembership{Member,~IEEE,}
	Zhaoqing Pan,~\IEEEmembership{Senior~Member,~IEEE,}
	Nam Ling,~\IEEEmembership{Life~Fellow,~IEEE}
	and~Qingming Huang,~\IEEEmembership{Fellow,~IEEE} 
	
	\thanks{\textcolor[rgb]{0,0,0}{D. Jin, J. Lei (Corresponding author), B. Peng, and Z. Pan are with the School of Electrical and Information Engineering, Tianjin University, Tianjin 300072, China (e-mail: jdc3159761141@tju.edu.cn; jjlei@tju.edu.cn;  bpeng@tju.edu.cn; zqpan3-c@my.cityu.edu.hk).}}
	\thanks{N. Ling is with the Department of Computer Science and Engineering, Santa Clara University, Santa Clara, CA 95053, USA (email: nling@scu.edu).}
	\thanks{Q. Huang is with the School of Computer Science and Technology, University of Chinese Academy of Sciences, Beijing 101408, China (e-mail: qmhuang@ucas.ac.cn).}
}
%

\maketitle

\begin{abstract}
2D image coding for machines (ICM) has achieved great success in coding efficiency, while less effort has been devoted to stereo image fields. To promote the efficiency of stereo image compression (SIC) and intelligent analysis, the stereo image coding for machines (SICM) is formulated and explored in this paper. More specifically, a machine vision-oriented stereo feature compression network (MVSFC-Net) is proposed for SICM, where the stereo visual features are effectively extracted, compressed, and transmitted for \textcolor[rgb]{0,0,0}{3D visual task}. To efficiently compress stereo visual features in MVSFC-Net, a stereo multi-scale feature compression  (SMFC) module is \textcolor[rgb]{0,0,0}{designed to gradually transform sparse stereo multi-scale features into compact joint visual representations by removing spatial, inter-view, and cross-scale redundancies simultaneously.} Experimental results show that the proposed MVSFC-Net obtains superior compression efficiency as well as 3D visual task performance, when compared with the existing ICM anchors recommended by MPEG and the state-of-the-art SIC method.
\end{abstract}

\begin{IEEEImpStatement}
With the rapid development of stereo image technologies and artificial intelligence, stereo images have been widely used in many fields, such as autonomous driving, smart city and intelligent industry. However, due to the inherent limitations of storage space and transmission bandwidth, the stereo images need to be efficiently compressed to better balance 3D visual analysis performance and transmission cost. In this paper, the stereo image coding for machines (SICM) is explored. Specifically, a novel machine vision-oriented stereo feature compression network (MVSFC-Net) is proposed to achieve high-efficiency SICM. In particular, a stereo multi-scale feature compression (SMFC) module is designed to gradually transform sparse stereo multi-scale features into compact joint visual representations by removing spatial, inter-view, and cross-scale redundancies simultaneously. It is hoped the proposed MVSFC-Net would benefit the development of the learned stereo image compression field and be helpful to promote 3D visual analysis services under the limited capabilities of stereo image storage and transmission.
\end{IEEEImpStatement}


\section{Introduction}

\IEEEPARstart{W}{ith} the rapid development of stereo image technologies and artificial intelligence, enormous intelligence applications (including Smart City, Autonomous Driving, and Intelligent Industry, etc) have sprouted to replace traditional man-powered pipelines \cite{2024-TAI-Stereo}, \cite{2024-TPAMI-VCM}, \cite{2022-TAI-Stereo}. To complete various three-dimensional (3D) visual tasks, these intelligence applications generally need to efficiently handle and analyze massive stereo images, resulting in a huge burden on the data transmission and storage system. Therefore, it is highly desirable to develop high-efficiency stereo image coding for machines (SICM).

Aiming to reconstruct high-quality stereo images under bitrate constraints, traditional stereo image coding (SIC) methods \cite{2016-3D-HEVC} have been developed over several decades, and the coding efficiency has been improved dramatically. These methods generally employed a traditional hybrid coding framework, where the signal redundancies are removed by multiple hand-crafted technologies, \textcolor[rgb]{0,0,0}{such as} intra-prediction, inter-view prediction, and entropy coding \cite{2024-TBC-Li}, \cite{2011-TBC-stereovideocoding}. To achieve end-to-end rate-distortion (RD) optimization, recent methods applied deep learning technologies into SIC to further boost compression efficiency \cite{2022-BCSIC}, \cite{2022-SASIC}.
Nevertheless, the aforementioned methods generally focused on reconstructing stereo images for human visual systems (HVS). Since pixel-level signal fidelity rather than high-level semantic fidelity is emphasized, these SIC methods are shown suboptimal 3D visual task performance in SICM, especially at low bitrates.


In recent years, 2D image coding for machines (ICM) \cite{2023-Acccess-Cao}, \cite{2020-TIP-VCM} has attracted more and more attention. 
\textcolor[rgb]{0,0,0}{From the perspective of compressing the source image or visual features}, existing ICM methods can be categorized into two pipelines, \textcolor[rgb]{0,0,0}{\textit{i.e.}, compress-then-analyze (CTA) \cite{2020-Mei}, \cite{2022-TCSVT-LSM}, and analyze-then-compress (ATC) \cite{2018-ICIP-Choi},\cite{2022-ECCV-Feng}.} As for the CTA pipeline, the \textcolor[rgb]{0,0,0}{source image is} compressed and reconstructed for  \textcolor[rgb]{0,0,0}{visual analysis}. To facilitate both compression efficiency and visual analysis performance, researchers have concentrated on reconstructing images that are more suitable for subsequent visual tasks, such as latent space masking \cite{2022-TCSVT-LSM}, joint training with visual loss \cite{2023-TMM-Gao}, \cite{2020-ICIP-SPIC},  \cite{2021-CAS-Wang} and so on. Different from the CTA pipeline \textcolor[rgb]{0,0,0}{compressing the source image}, the ATC pipeline focuses on compressing visual features which are expected to have lower information entropy than the whole high-quality images. In specific, representative visual features \cite{2018-ICIP-Choi}, \cite{2022-ECCV-Feng}  are extracted at the front-end, and then compressed as well as transmitted to the service-end for visual analysis. As such, the ATC pipeline can effectively reduce the sizes of bit-stream and decoding computational complexity. However, existing methods mainly focus on ICM, unprecedented challenges remain in the exploration of SICM. 

In contrast to 2D images, stereo images enable a dynamic scene visualization with more viewpoints, which leads to a significant increase in the data volume \cite{2024-TBC-Zhang}, \cite{2024-TAI-Liu}, \cite{2024-TAI-Kang}. Apart from removing the spatial redundancies \textcolor[rgb]{0,0,0}{as the ICM}, the inter-view redundancies between different viewpoints should also be considered \textcolor[rgb]{0,0,0}{in the SICM} to further improve compression efficiency and 3D visual analysis performance. 
As a result, applying the aforementioned ICM approaches directly to SICM struggles to achieve superior performance, due to no explicit exploration of the inter-view correlations between viewpoints.


To this end, this paper proposed a machine vision-oriented stereo feature compression network (MVSFC-Net) for SICM. To ensure the performance of both compression and  visual analysis, the stereo visual features are efficiently extracted, compressed, and transmitted in the MVSFC-Net. Generally, the stereo visual features own lower information entropy, while having higher-dimensional data size. Therefore, the key in the MVSFC-Net lies in the efficient redundancies removal among these sparse stereo features. Thus, a stereo multi-scale feature compression (SMFC) module is proposed, where the compact joint visual representation of the stereo features is progressively generated for \textcolor[rgb]{0,0,0}{higher compression efficiency.} The contributions are summarized as follows. 

\begin{itemize}
	\item \textcolor[rgb]{0,0,0}{To promote stereo image compression efficiency and 3D visual analysis performance, a machine vision-oriented stereo feature compression network (MVSFC-Net) is proposed. As far as we know, the proposed MVSFC-Net is the first exploration of the SICM.}
	
	\item A stereo multi-scale feature compression (SMFC) module is proposed for the stereo visual features compression, in which the intra, inter-view, as well as cross-scale redundancies are simultaneously removed to obtain \textcolor[rgb]{0,0,0}{compact joint visual representation for high-effieiceny compression}.
	\item Experimental results prove that the proposed  MVSFC-Net outperforms \textcolor[rgb]{0,0,0}{the existing ICM anchors recommended by MPEG and the state-of-the-art SIC method} by a large margin in terms of the compression efficiency for 3D visual task performance. 
\end{itemize}

The remainder of the article is organized as follows. Section \uppercase\expandafter{\romannumeral2} systematically reviews  the related works on SIC and ICM. After that, Section \uppercase\expandafter{\romannumeral3} formulates the SICM and introduces the proposed MVSFC-Net for SICM in detail.  In  Section \uppercase\expandafter{\romannumeral4}, experimental results and analysis are reported. Finally, Section \uppercase\expandafter{\romannumeral5}  concludes this paper.

\section{Related Work}\label{section2}



\subsection{Stereo Image Compression for HVS} 

During the past decades, with the aim to reconstruct high-fidelity stereo images for HVS with the restriction of bitrate, \textcolor[rgb]{0,0,0}{SIC has been widely studied for efficient stereo image storage and transmission \cite{2002-TCSVT-SIC}, \cite{2009-TCSVT-SIC}, \cite{2012-TCSVT-SVC}.}
Different from 2D image compression, the SIC generally focuses on removing inter-view redundancies by exploiting the correlation between stereo images. For instance, on the basis of high efficiency video coding (HEVC), the multiview HEVC (MV-HEVC) \cite{2011-IEEE-MVC} significantly improves the SIC performance by extending multi-view coding technologies, such as disparity-compensated prediction (DCP) \cite{2024-TCSVT-Zhang}. To improve the DCP accuracy,  Wong et al.  \cite{2012-TCSVT-SVC} considered the horizontal scaling and shearing (HSS) deformations in stereo images and introduced HSS-based DCP to improve stereo image compression efficiency. 	

With the growing success of deep learning, several researchers have exploited neural networks to construct the learned stereo image compression framework. Compared to traditional methods, learned stereo image compression methods can boost the compression efficiency \textcolor[rgb]{0,0,0}{by jointly optimizing the whole framework with the rate-distortion cost}. For instance,  Lei et al. \cite{2022-BCSIC} employed bi-directional coding mechanism in stereo image compression and achieved impressive coding performance improvement. Wodlinger et al.  \cite{2022-SASIC} proposed stereo image compression with latent shifts and stereo attention (SASIC), where the encoded left-image latent representation is shifted to the \textcolor[rgb]{0,0,0}{right-image as an inter-view reference prior}.

Despite significant progress on SIC for human perceptual, it is undesirable to directly apply the above methods for SICM, due to the emphasized pixel-level signal fidelity rather than high-level semantic fidelity for visual analysis.

\begin{figure*}[htbp]
	\centering
	\includegraphics[width =18.0cm]{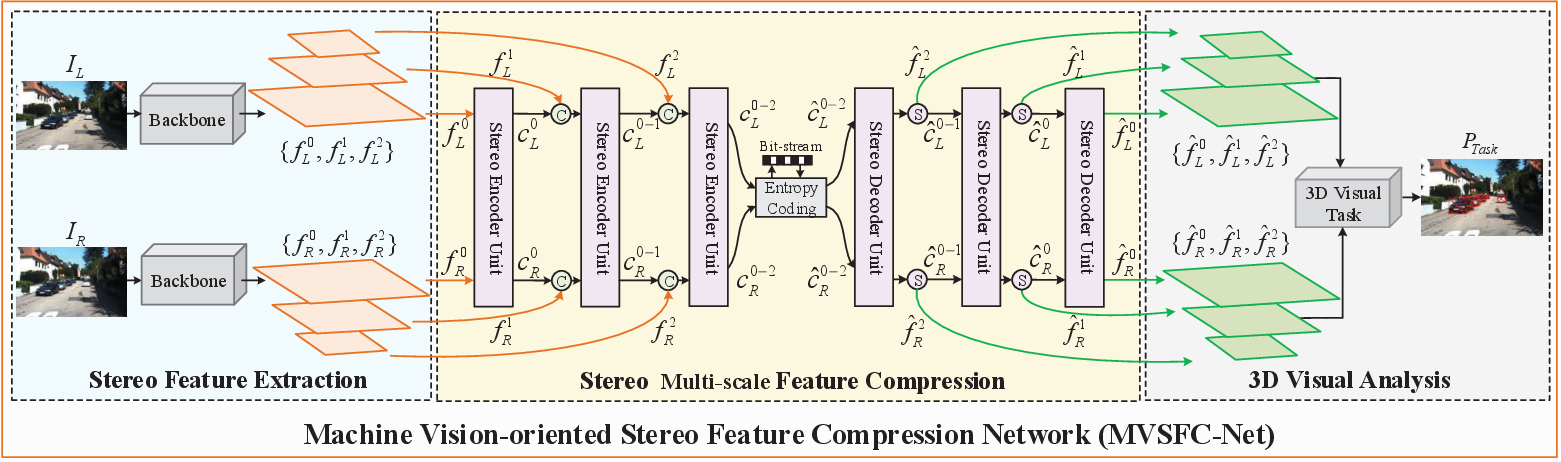}
	\caption{\textcolor[rgb]{0,0,0}{The architecture of the proposed MVSFC-Net. For the encoding stereo images $ \left\{I_{L}, I_{R}\right\} $, the stereo feature extraction module is firstly applied to obtain the stereo multi-scale features $ \left\{f_{L}^{i}, f_{R}^{i}\|i\in 0,1,2\right\} $. Then, the $ \left\{f_{L}^{i}, f_{R}^{i}\|i\in 0,1,2\right\} $ are efficiently compressed by the proposed stereo multi-scale feature compression module. Finally, the visual analysis module deployed at service-end is utilized to perform vision task based on reconstructed stereo multi-scale features $ \left\{\hat{f}_{L}^{i}, \hat{f}_{R}^{i}\|i\in 0,1,2\right\} $}.}
	
	\label{fig1}
\end{figure*}

\subsection{Image Coding for Machines}


\textcolor[rgb]{0,0,0}{In the era of deep learning, to promote image compression efficiency for visual tasks, many efforts have been devoted to ICM based on CTA pipeline and ATC pipeline.}

As for the CTA pipeline, the source image is compressed at the front-end and reconstructed at the decoder side for visual analysis. \textcolor[rgb]{0,0,0}{To reduce bitrate while maintaining visual task performance}, several researchers focused on masking out information that is not required for the visual task. For instance, Fischer et al. \cite{2022-TCSVT-LSM} proposed a latent space masking network (LSMnet) to discriminatively squeeze out the non-salient components in the compact latent representation. Another promising solution is to enhance the image coding framework with the visual task-related loss optimization. Concretely, Fischer et al. \cite{2022-MMSP-FRDO} replaced conventional pixel-level rate-distortion optimization (RDO) with a standard-compliant feature-based RDO (FRDO) for VVC to improve the object detection accuracy. \textcolor[rgb]{0,0,0}{Gao et al. \cite{2023-TMM-Gao} proposed semantics-oriented metrics to encourage the reconstructed images more suitable for various visual tasks.} Patwa et al. \cite{2020-ICIP-SPIC} incorporated a cross-entropy loss term in the end-to-end image compression framework to improve the image classification performance.  Wang et al. \cite{2021-CAS-Wang} proposed to optimize the image compression network with the combination of visual task loss and the rate-distortion loss. \textcolor[rgb]{0,0,0}{Although these methods obviously improve the visual task performance under bitrate constraints for CTA pipeline, redundancy information for machines in the reconstructed color images is inevitably maintained, which prevents further improvements for coding performance.}

In the ATC pipeline, the compact features are firstly extracted at the front-end, and then encoded as well as transmitted to the service end for visual analysis. In particular, MPEG has designed a series of visual feature descriptors \cite{2018-TMM-CDVS-IPS}, \cite{2019-TMM-CDVS-CBF} in CDVS \cite{2016-TIP-CDVS} for visual search and video analysis. Choi et al. \cite{{2018-ICIP-Choi}} developed a collaborative intelligence for mobile applications, where the deep features are near-lossless and lossy compressed for object detection. To enhance the feature generalization for multiple visual tasks, Feng et al. \cite{2022-ECCV-Feng} proposed to obtain omnipotent features by taking advantage of self-supervised learning technologies. \textcolor[rgb]{0,0,0}{Liu et al. \cite{2023-ICASSP-Liu} proposed an efficient feature compression network, which explores the correlation between features of different scales, and utilizes the encoded small-scale features to predict large-scale features.} Sun et al. \cite{2021-TCSVT-Sun} proposed to separately represent the object in a semantically structured bitstream, where visual tasks are performed on the partial bitstream to reduce the decoding computational complexity and the transferred data. To fulfill the needs for machine vision and human vision jointly, several efforts have been devoted to developing a scalable coding framework \cite{2021-TMM-Yang-SFIC}, \cite{2022-TMM-Wang},  \cite{2022-TCSVT-Huang}, where an additional bitstream is designed to ensure signal fidelity for HVS. For instance, Yang et al. \cite{2021-TMM-Yang-SFIC} proposed a face image coding framework, where the compact structure representations are extracted to be the basic layer for machine vision, and the color representation is served as an enhanced layer to generate images for human vision. Wang et al. \cite{2022-TMM-Wang} proposed a collaborative visual information representation to investigate interactions between the visual feature for machine vision and the texture for human vision. Huang et al. \cite{2022-TCSVT-Huang} proposed an efficient human-machine friendly video coding scheme (HMFVC), where the semantic information for machine vision is efficiently exploited to enhance the video reconstruction quality for human vision.

However, as far as we know, no previous research  on SICM has been reported. Since the inter-view correlations between two viewpoints have not been explicitly explored, it is not satisfactory to apply the aforementioned \textcolor[rgb]{0,0,0}{ICM methods} for SICM.

	\section{The Proposed Method}\label{section3}

\textcolor[rgb]{0,0,0}{In this section, we elaborate the proposed method for SICM in detail. First, the formulations of SICM and the overall framework of machine vision-oriented stereo feature compression network are systematically introduced. Second, the proposed stereo multi-scale feature compression module is illustrated.} Finally, the loss functions and implementation details are described.

\subsection{ Formulations of SICM and Framework of MVSFC-Net}

\textcolor[rgb]{0,0,0}{To promote the stereo image compression efficiency torwards 3D machine vision  rather than human vision, the stereo image compression for machines is formulated and researched in this paper.} Formally, the SICM aims to maximize the machine task under the constraints on the bitrate, the formulations can be summarized as follows: 

\begin{equation}\label{qua1}
	\arg\max\limits_{\theta}{P_{task}, \textit{s.t.} \sum_{i=1}^{n}R_{i} \leq R_{max}},
\end{equation}
where $ P_{task} $ denotes visual task performance, $ R_{i} $ is the  bitrate consumption of a viewpoint, $ n $ represents the number of viewpoints, $ R_{max} $  denotes the bitrate limit, and $ \theta $ represents the parameters of the network architecture. 

\textcolor[rgb]{0,0,0}{To solve the above SICM problem, the MVSFC-Net is correspondingly designed in this paper.} In the MVSFC-Net, 3D object detection is employed as the 3D visual task, as it is an important research topic in computer vision \cite{2022-TPAMI-Qin}, \cite{2018-TPAMI-Chen}, \cite{2021-ICRA-YOLOStereo3D}. Fig. \ref{fig1}  illustrates the framework of MVSFC-Net. As shown in the figure, the MVSFC-Net is composed of \textcolor[rgb]{0,0,0}{stereo feature extraction module, SMFC module, and visual analysis module.} Specifically, the stereo feature extraction module is utilized to extract representative visual features, which are \textcolor[rgb]{0,0,0}{capable enough to represent the 3D scene characteristics for the 3D object detection task, while having less information entropy than the full-resolution stereo image.} Generally, the high-resolution feature maps from earlier layers are supposed to preserve more original details (\textit{i.e.}, texture and contours of the object), while low-resolution feature maps from deeper layers can capture semantic and contexture information. Accordingly, 
\textcolor[rgb]{0,0,0}{to ensure satisfactory  machine task performance, the stereo multi-scale feature $ \left\{f_{L}^{i}, f_{R}^{i}\|i\in 0,1,2\right\} $ from different layers in the stereo feature extraction module are served as encoding information in the MVSFC-Net.  \textcolor[rgb]{0,0,0}{After the stereo feature extraction module}, the SMFC module is designed to efficiently compress $ \left\{f_{L}^{i}, f_{R}^{i}\|i\in 0,1,2\right\} $ by exploiting complex correspondences among these features. Finally, the visual analysis module is employed to produce the 3D object detection result based on the  reconstructed stereo multi-scale features $ \left\{\hat{f}_{L}^{i}, \hat{f}_{R}^{i}\|i\in 0,1,2\right\} $.} Owing to compressing and transmitting representative visual features rather than pixel-level source stereo image, \textcolor[rgb]{0,0,0}{the proposed MVSFC-Net could obtain higher detection accuracy with lower bitrate consumption  and computational complexity.}

\subsection{Stereo Multi-scale Feature Compression Module}


Stereo multi-scale features have higher-dimensional data sizes but less information entropy, thus effectively removing redundancies among these sparse features is crucial for MVSFC-Net. Therefore, the SMFC module is proposed to efficiently compress the stereo multi-scale features, where the spatial, inter-view, and cross-scale redundancies are removed simultaneously.


The architecture of the SMFC module is illustrated in Fig. \ref{fig1}. As shown in the figure, taking the sparse stereo multi-scale features $ \left\{f_{L}^{i}, f_{R}^{i}\|i\in 0,1,2\right\} $ as input, the  SMFC module firstly converges the representative information among multi-scale features and generates the compact joint \textcolor[rgb]{0,0,0}{visual } representation $ \left\{c_L^{0-2},c_R^{0-2}\right\} $. After that, the $ \left\{c_L^{0-2},c_R^{0-2}\right\} $ are  entropy encoded into bit-stream, which is then decoded in the decoder side to progressively recover the reconstructed stereo multi-scale features $ \left\{\hat{f}_{L}^{i}, \hat{f}_{R}^{i}\|i\in 0,1,2\right\} $ for further visual analysis.

At the beginning of the SMFC module, the first \textcolor[rgb]{0,0,0}{low-level} stereo feature $ \left\{f_L^0,f_R^0\right\} $ are fed into a stereo encoder unit to obtain a \textcolor[rgb]{0,0,0}{low-level} compact \textcolor[rgb]{0,0,0}{visual} representation $ \left\{c_L^0,c_R^0\right\} $,\textit{ i.e.},
\begin{equation}\label{qua2}
\left\{c_L^0,c_R^0\right\} = SEU(f_L^0,f_R^0),
\end{equation}
where \textcolor[rgb]{0,0,0}{$ SEU(\cdot) $} denotes the stereo encoder unit, which is composed of two downsampling convolutions and a bi-directional contextual transform \cite{2022-BCSIC}. \textcolor[rgb]{0,0,0}{In particular, the downsampling convolution is applied to reduce the spatial resolution for removing spatial redundancies, while the bi-directional contextual transform is utilized to remove inter-view redundancies by interacting mutual information between $ \left\{f_L^0,f_R^0\right\} $.}
Since the outputs of different layers still describe the same dynamic scenes, the medium-level stereo features $ \left\{f_L^1,f_R^1\right\} $ are jointly compressed with the low-level compact representation $ \left\{c_L^0,c_R^0\right\} $ to remove cross-scale redundancies, and the  compact joint \textcolor[rgb]{0,0,0}{visual} representation of low-level and medium-level features $ \left\{c_L^{0-1},c_R^{0-1}\right\} $ are generated,\textit{ i.e.}, 
\begin{equation}\label{qua2}
\left\{c_L^{0-1},c_R^{0-1}\right\} = SEU(c_L^0 \oplus f_L^1,c_R^0 \oplus f_R^1 ),
\end{equation}
where $\oplus$ denotes the channel-wise concatenation. Following that, the high-level stereo feature $ \left\{f_L^2,f_R^2\right\} $  are \textcolor[rgb]{0,0,0}{merged} with $ \left\{c_L^{0-1},c_R^{0-1}\right\} $ to obtain final compact joint \textcolor[rgb]{0,0,0}{visual} representations $ \left\{c_L^{0-2},c_R^{0-2}\right\} $, which are then entropy encoded into bit-stream. 

The decoder side of the SFMC module is symmetric with the encoder side. The reconstructed stereo multi-scale features $ \left\{\hat{f}_{L}^{i}, \hat{f}_{R}^{i}\|i\in 0,1,2\right\} $ are sequentially decoded from the reconstructed compact joint \textcolor[rgb]{0,0,0}{visual} representation,\textit{ i.e.},
\begin{equation}\label{qua2}
\left\{\hat{c}_L^{0-1}, \hat{c}_R^{0-1}\right\}, \left\{\hat{f}_{L}^{2},\hat{f}_{R}^{2} \right\} = Split(SDU(\hat{c}_L^{0-2}, \hat{c}_R^{0-2})), 
\end{equation}
\begin{equation}\label{qua3}
\left\{\hat{c}_{L}^{0}, \hat{c}_R^{0}\right\}, \left\{\hat{f}_{L}^{1},\hat{f}_{R}^{1} \right\} = Split(SDU(\hat{c}_L^{0-1}, \hat{c}_R^{0-1})), 
\end{equation}
\begin{equation}\label{qua4}
\left\{\hat{f}_{L}^{0},\hat{f}_{R}^{0} \right\} = SDU(\hat{c}_L^{0}, \hat{c}_R^{0}), 
\end{equation}
where $ Split\left(\cdot\right) $ is the channel-wise split, and $ SDU(\cdot) $ denotes the stereo decoder unit, which is composed of a bi-directional contextual transform module \cite{2022-BCSIC} and two transposed convolutions.  

In contrast to separately dealing with stereo multi-scale features $ \left\{f_{L}^{i}, f_{R}^{i}\|i\in 0,1,2\right\} $, the SMFC progressively transforms the sparse stereo multi-scale features into the compact joint \textcolor[rgb]{0,0,0}{visual} representation $ \left\{c_L^{0-2},c_R^{0-2}\right\} $, in which the spatial, inter-view as well as cross-scale redundancies are simultaneously reduced to improve the compression efficiency.

\subsection{Loss Functions and Implementation Details}

The proposed MVSFC-Net is trained in an end-to-end manner by the rate-distortion function, which comprises a task distortion term and a bitrate consumption term. To achieve optimal compression performance for the machine vision task, the task distortion term is calculated by the visual task loss, with the aim to reconstruct high-quality stereo multi-scale features for visual analysis. More specifically, we follow the method in \cite{2021-ICRA-YOLOStereo3D} to calculate the visual task loss, which consists of three parts, including a focal loss for anchor classification $ \mathcal{L}_{cls} $, a smooth L1 loss for box regression  $ \mathcal{L}_{reg} $, and a stereo focal loss on disparity  $ \mathcal{L}_{dis} $, \textit{i.e.}, 
\begin{equation}\label{equa7}
D_v =  \mathcal{L}_{cls} + \mathcal{L}_{reg} + \mathcal{L}_{dis}.
\end{equation}
Finally, the overall rate-distortion loss function can be summarized as follow:
\begin{equation}\label{equa7}
\mathcal{L}_{T} = \lambda D_v + R_l + R_r,
\end{equation}
where the $  R_l $ and $ R_r $ denote the bitrate of the left image and the right image, respectively, and $ \lambda $ is the parameter to tradeoff the rate-distortion cost, which is set to $ \left\{0.5, 1, 4, 16, 64, 256\right\} $. In terms of the detailed architecture of the MVSFC-Net, the stereo feature extraction module and visual analysis module are the same as the \cite{2021-ICRA-YOLOStereo3D}. More specifically, the backbone in the stereo feature extraction module is implemented by ResNet-34 \cite{ResNet-34}. The visual analysis module consists of multi-stage fusion, box regression, and disparity estimation to obtain the final 3D object detection result.

\begin{table*}[!t]
\renewcommand{\arraystretch}{2.0}
\scriptsize
\centering
\caption{\textcolor[rgb]{0,0,0}{Comparison Results in terms of BD-rate and BD-$ {\rm AP_{3D}} $}}
\setlength{\tabcolsep}{0.5mm}{
	\begin{tabular}
		{p{1.3cm}<{\centering}|cc|cc|cc|cc|cc|ccccc}
		\hline
		\hline
		\multirow {2}{*}{\textbf{Categories}} & \multicolumn{2}{c|}{\textbf{VVC-Feature \cite{VVC-Feature}}} &
		\multicolumn{2}{c|}{\textbf{VVC-Inter  \cite{Image-anchor}}} &
		\multicolumn{2}{c|}{\textbf{Liu's method \cite{2023-ICASSP-Liu}}} &
		\multicolumn{2}{c|}{\textbf{Wang's method \cite{2021-CAS-Wang}}} &
		\multicolumn{2}{c|}{\textbf{Lei's method \cite{2022-BCSIC}}} &
		\multicolumn{2}{c}{\textbf{Proposed}}\\
		\cline{2-13}
		
		& \textbf{ \ \ BD-rate \ \ } & \textbf{BD-$ \mathbf {AP_{3D}} $} & \textbf{\ \ BD-rate \ \ } &\textbf{BD-$ \mathbf {AP_{3D}} $ } & \textbf{\ \ BD-rate \ \ } & \textbf{BD-$ \mathbf {AP_{3D}} $} & \textbf{\ \ BD-rate \ \ } & \textbf{BD-$ \mathbf {AP_{3D}} $} & \textbf{\ \ BD-rate \ \ } & \textbf{BD-$ \mathbf {AP_{3D}} $} &  \textbf{\ \ BD-rate\ \ } & \textbf{BD-$ \mathbf {AP_{3D}} $}\\	
		
		\cline{1-13}
		\textbf{Car}
		& 418.728\% & -24.685\%  &	-28.462\% & 3.202\% &	-61.022\% & 6.971\% & -52.876\% & 7.834\% & -58.420\% &	7.796\% & \textbf{-78.254\%} & \textbf{12.477\%} \\ 
		\cline{1-13}
		\textbf{Pedestrian}
		&960.261\% & -16.774\% & -25.906\% & 1.344\% &	-38.717\% & 2.461\% & -49.041\% & 2.704\% & -49.225\% & 3.165\% & \textbf{-83.756\%}  &\textbf{6.259\%} \\
		\cline{1-13}
		\textbf{AVG}
		&689.495\% & -20.730\%  & -27.184\% & 2.273\% &	-49.870\% & 4.716\% & -50.959\% & 5.269\% & -53.822\% &	5.480\% &	\textbf{-81.005\%}  & \textbf{9.368\%} \\
		\hline
		\hline	
\end{tabular}}%
\label{table1}%
\end{table*}%

\begin{table*}[!t]
\renewcommand{\arraystretch}{2.0}
\scriptsize
\centering
\caption{ \textcolor[rgb]{0,0,0}{Comparison Results in terms of BD-rate and BD-$ {\rm AP_{BEV}} $}}
\setlength{\tabcolsep}{0mm}{
	\begin{tabular}
		{p{1.2cm}<{\centering}|cc|cc|cc|cc|cc|ccccc}
		\hline
		\hline
		\multirow {2}{*}{\textbf{Categories}} & \multicolumn{2}{c|}{\textbf{VVC-Feature \cite{VVC-Feature}}} &
		\multicolumn{2}{c|}{\textbf{\textcolor[rgb]{0,0,0}{VVC-Inter}  \cite{Image-anchor}}} &
		\multicolumn{2}{c|}{\textbf{Liu's method \cite{2023-ICASSP-Liu}}} &
		\multicolumn{2}{c|}{\textbf{Wang's method \cite{2021-CAS-Wang}}} &
		\multicolumn{2}{c|}{\textbf{Lei's method \cite{2022-BCSIC}}} &
		\multicolumn{2}{c}{\textbf{Proposed}}\\
		\cline{2-13}
		
		& \textbf{ \ \ BD-rate \ \ } & \textbf{BD-$ \mathbf {AP_{BEV}} $} & \textbf{\ \ BD-rate \ \ } &\textbf{BD-$ \mathbf {AP_{BEV}} $ } & \textbf{\ \ BD-rate \ \ } & \textbf{BD-$ \mathbf {AP_{BEV}} $}  & \textbf{\ \ BD-rate \ \ } & \textbf{BD-$ \mathbf {AP_{BEV}} $} & \textbf{\ \ BD-rate \ \ } & \textbf{BD-$ \mathbf {AP_{BEV}} $} &  \textbf{\ \ BD-rate\ \ } & \textbf{BD-$ \mathbf {AP_{BEV}} $}\\	
		
		\cline{1-13}
		\textbf{Car}
		& 352.770\% &	-26.181\%  &	-29.061\% & 3.699\% &	-60.302\% & 7.368\% & -53.924\% & 9.347\%  & -62.560\% & 8.694\% & \textbf{-76.347\%} & \textbf{13.174\%} \\ 
		\cline{1-13}
		\textbf{Pedestrian}
		& 890.801\% & -18.384\% &   -26.190  \% &1.043\% &	-40.455\% & 2.378\% & -55.558\% & 3.359\% & -51.369\% & 3.503\% & \textbf{-77.472\%}  &\textbf{ 6.709\%} \\
		\cline{1-13}
		\textbf{AVG}
		&621.785\% & -22.282\% & -27.626\% & 2.371\% &	-50.378\% & 4.873\% & -54.741\% & 6.353\% & -56.965\% &	6.098\% &	\textbf{-76.910\%}  & \textbf{9.942\%} \\
		\hline
		\hline
\end{tabular}}%
\label{table2}%
\end{table*}%

	\begin{figure*}[htbp]

\centering
\includegraphics[width =16.7cm]{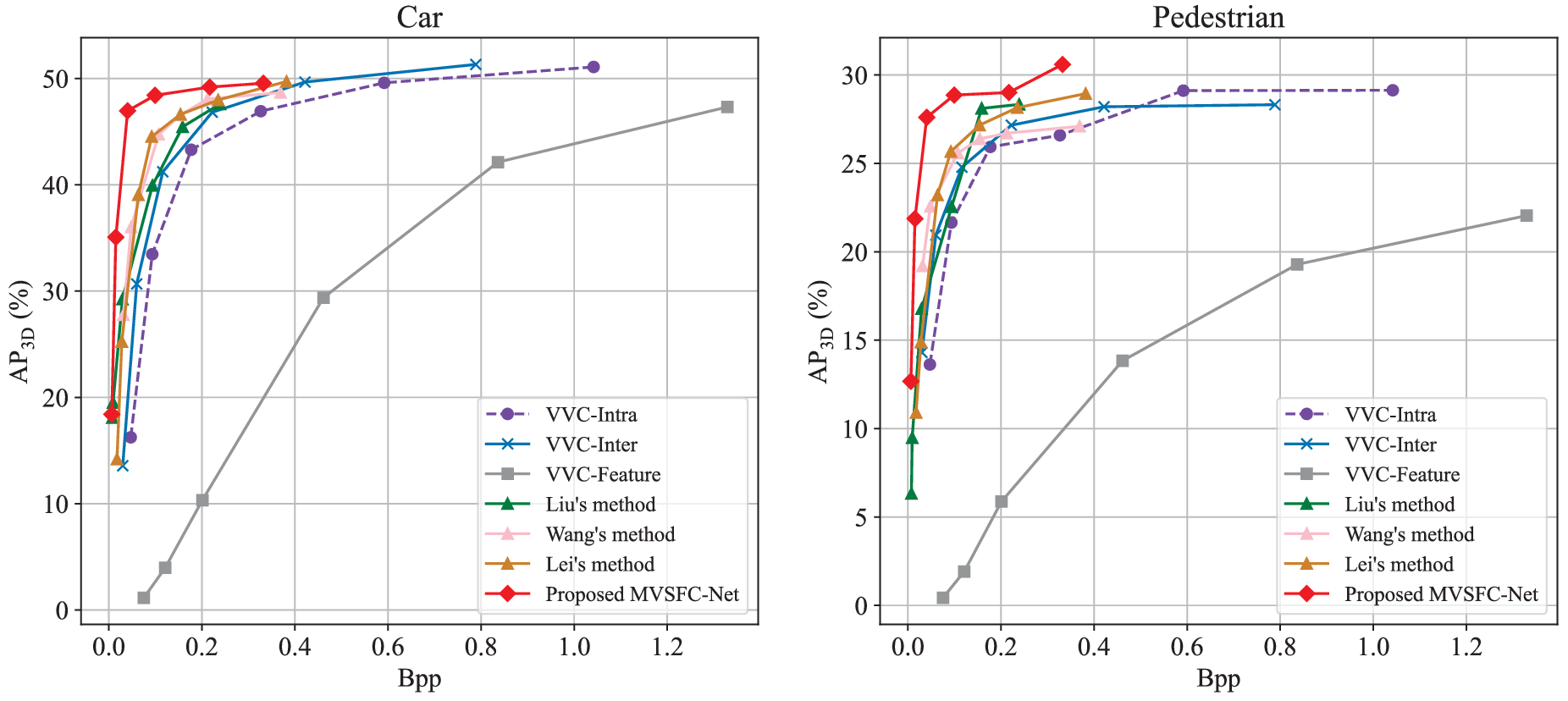}
\caption{\textcolor[rgb]{0,0,0}{Rate-distortion curves comparison when the distortion is measured by  $ \rm AP_{3D} $. }}
\label{fig2}
\end{figure*}


	\section{Experiments}\label{section4}
\subsection{Experimental Configurations}

\subsubsection{Datasets}
The MVSFC-Net is trained on the KITTI Object Detection Benchmark \cite{2012-CVPR-Kitti}, in which the training set and validation set are split by the Chen’s split \cite{2015-CVPR-Chen}. \textcolor[rgb]{0,0,0}{The top 100 pixels in the stereo images are cropped, and then the cropped stereo images are scaled to the resolution 288$\times$1280 for fast inference and training. }

\subsubsection{Training Strategy}
As the MVSFC-Net comprises several learnable components, it is hard to optimize the network stably by training from scratch. Therefore, a progressive training strategy is utilized to train the proposed MVSFC-Net. First,  the stereo feature extraction module and visual analysis module are initialized by a YOLOStereo3D model, which is pre-trained on the KITTI training set \cite{2012-CVPR-Kitti} for 80 epochs. Then, the MVSFC-Net is optimized for 20 epochs by the loss $ \mathcal{L}_{T} $, with the parameters in the stereo feature extraction module and the visual analysis module fixed. Finally, all components in MVSFC-Net are end-to-end optimized together with the loss $ \mathcal{L}_{T} $ for the final 80 epochs, which is helpful to improve the compression efficiency for visual tasks. In summary, the MVSFC-Net is optimized using Adam \cite{2014-Adam} with batch-size 4 for 180 epochs (\textit{i.e.}, 80+20+80), where the learning rate is set as $ 1\times10^{-4} $, and decayed to $ 5\times10^{-6} $ by the CosineAnnealingLR. The proposed MVSFC-Net is implemented with Pytorch and trained on a PC with Inter i9-12900K CPU @ 3.20 GHz and a GeForce GTX 3090Ti GPU.

\subsection{Comparison Results and Analysis}

\subsubsection{Comparison Algorithms and Evaluation Metrics}

\begin{figure*}[htbp]
	\centering
	\includegraphics[width =16.7cm]{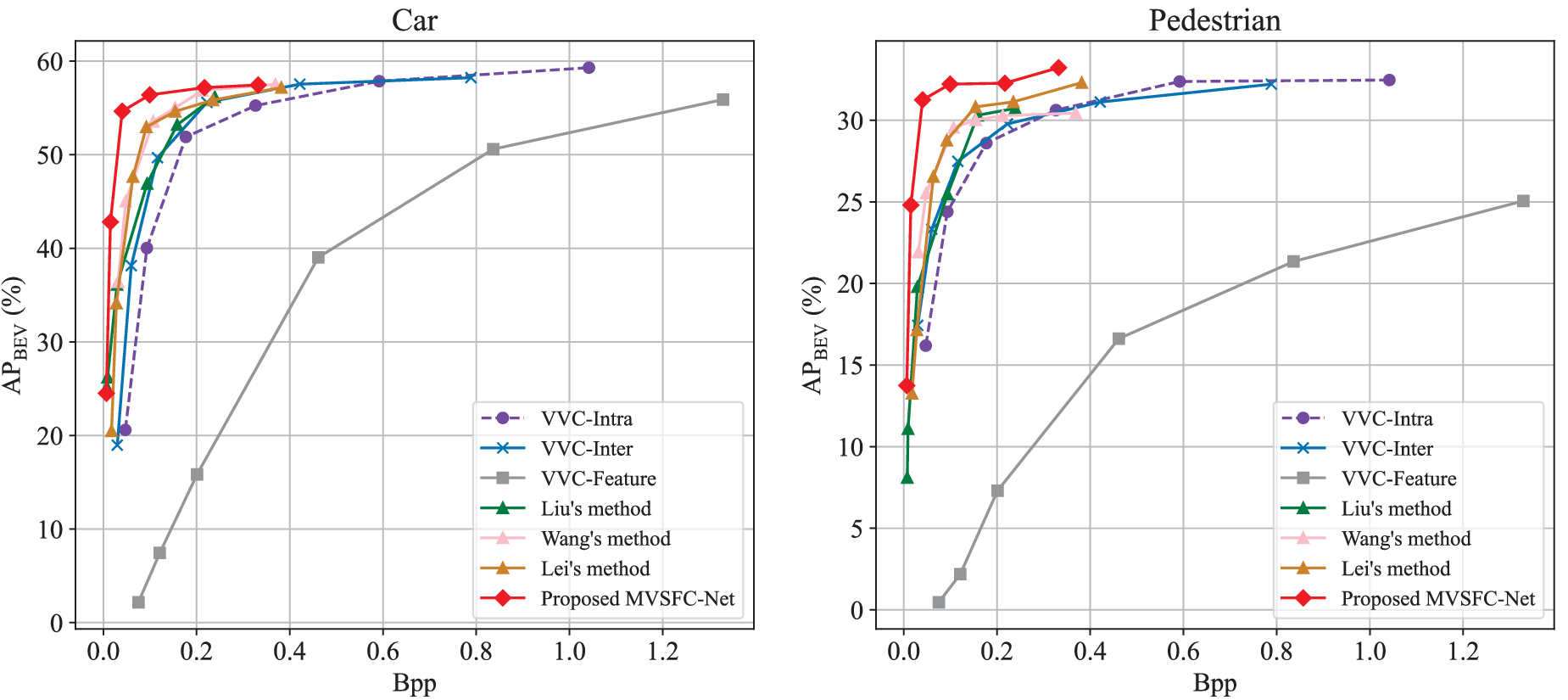}
	\caption{\textcolor[rgb]{0,0,0}{Rate-distortion curves comparison when the distortion is measured by  $ \rm AP_{BEV} $.}}
	\label{fig3}
\end{figure*}
To verify the efficiency of the proposed MVSFC-Net, ICM anchors recommended by MPEG (\textit{i.e.}, VVC image anchor \cite{Image-anchor} and VVC feature anchor \cite{VVC-Feature}), \textcolor[rgb]{0,0,0}{ICM algorithm proposed in Wang's method \cite{2021-CAS-Wang}}, \textcolor[rgb]{0,0,0}{Liu's method \cite{2023-ICASSP-Liu}}, and state-of-the-art learned stereo image compression algorithm proposed in Lei's method \cite{2022-BCSIC} are utilized for comparison. As for the evaluation of the  Wang's method \cite{2021-CAS-Wang}, Lei's method \cite{2022-BCSIC}, and VVC image anchor \cite{Image-anchor},
the whole stereo images are firstly compressed into bit-stream to calculate bitrate, and then the reconstructed image is input into the YOLOStereo3D model \cite{2021-ICRA-YOLOStereo3D} to evaluate the performance of 3D detection task. For a fair comparison, the Wang's method \cite{2021-CAS-Wang}, \textcolor[rgb]{0,0,0}{Liu's method \cite{2023-ICASSP-Liu},} and Lei's method \cite{2022-BCSIC} are retrained on the same KITTI training set as the proposed method. It is worthwhile to mention that both All-Intra configuration \footnote{EncoderApp.exe -c encoder\_intra\_vtm.cfg -i input.yuv -b output.bin -o output.yuv -wdt 1248 -hgt 376 -q QP --InputChromaFormat=420} and Low Delay P (LDP) configuration \footnote{EncoderApp.exe -c encoder\_lowdelay\_P\_vtm.cfg -i input.yuv -b output.bin -o output.yuv -wdt 1248 -hgt 376 -q QP InputChromaFormat=420 \\  \indent QP is set to 22, 27, 32, 37, 42, and 47.} under JVET common test conditions (CTC) are tested for VVC image anchor \cite{Image-anchor}, \textcolor[rgb]{0,0,0}{denoted as VVC-Intra and VVC-Inter}, respectively. More specifically, the left image is compressed firstly and then the right image is compressed upon the compressed left image in the LDP configuration, while the left image and right image are compressed dependently in the All-Intra configuration. 
In this paper, the performance of the VVC feature anchor \cite{VVC-Feature}, denoted as VVC-Feature, is evaluated by following the method suggested in \cite{VVC-Feature}. In the VVC-Feature, the stereo multi-scale features are pre-processed, quantized, packed, and compressed by the VVC reference software VTM-12.0 \cite{VTM12.20}, and the reconstructed features are sent to the visual evaluation to measure the 3D visual performance. 

Regarding the evaluation metric, the widely adopted BD-rate \cite{RD} and BD-AP \cite{2022-TCSVT-LSM} are  calculated to measure the coding performance. Specifically, the distortion is calculated by the accuracy of 3D visual task performance, including average precision for 3D box ($ \rm AP_{3D} $) and bird’s eye view ($ \rm AP_{BEV} $). The bitrate is measured by the bits per pixels (BPP), which is calculated by the coding bits divided by the total number of pixels in the original images. In this paper, the VVC-Intra is set as the anchor to calculate the BD-rate and BD-AP.

\subsubsection{Coding Performance}
Table \ref{table1} presents the BD-rate as well as the BD-AP comparison results in terms of the $ \rm AP_{3D} $. The positive value of the BD-rate indicates a rise in bit-rate consumption when compared to the anchor, while the negative value indicates a bit-rate saving. As shown in the table, the MVSFC-Net obtains an average of 81.005\% BD-rate reduction against the anchor. In comparion, the Lei's method \cite{2022-BCSIC}, Wang's method \cite{2021-CAS-Wang}, \textcolor[rgb]{0,0,0}{Liu's method \cite{2023-ICASSP-Liu}, }and VVC-Inter \cite{Image-anchor} achieve 53.822\%, 50.959\%, \textcolor[rgb]{0,0,0}{49.870\%, }and 27.184\% BD-rate reduction respectively, and the VVC-Feature \cite{VVC-Feature} brings obvious BD-rate increase. Overall, the proposed MVSFC-Net achieves higher performance than other comparison methods by a large margin, which strongly demonstrates the huge potential of SICM and the effectiveness of the proposed MVSFC-Net.  
Consistent results are observed when the  distortion is measured by the  $ \rm {AP_{BEV}} $, and the BD-rate as well as BD-$\rm {AP_{BEV}} $ comparison results are presented in Table \ref{table2}. As shown in the table, the proposed MVSFC-Net also achieves the best compression efficiency among all comparison methods with an average of 76.910\% BD-rate reduction as well as 9.942\% $ \rm {AP_{BEV}} $ improvement. Overall, the proposed MVSFC-Net shows the best compression efficiency in SICM scenarios where machine task matters.

Fig. \ref{fig2} and Fig. \ref{fig3} show the rate-distortion curves comparison results when the distortion is measured by $ \rm AP_{3D} $ and $ \rm AP_{BEV} $, respectively. It can be observed that the proposed MVSFC-Net behaves better than all comparison methods on all bitrates. It is worth mentioning that the proposed MVSFC-Net provides greater gains at lower bitrate, as the MVSFC-Net only needs to compress feature-domain compact representation with lower information entropy.

\begin{figure*}[htbp]
	\centering
	\includegraphics[width =17.7cm]{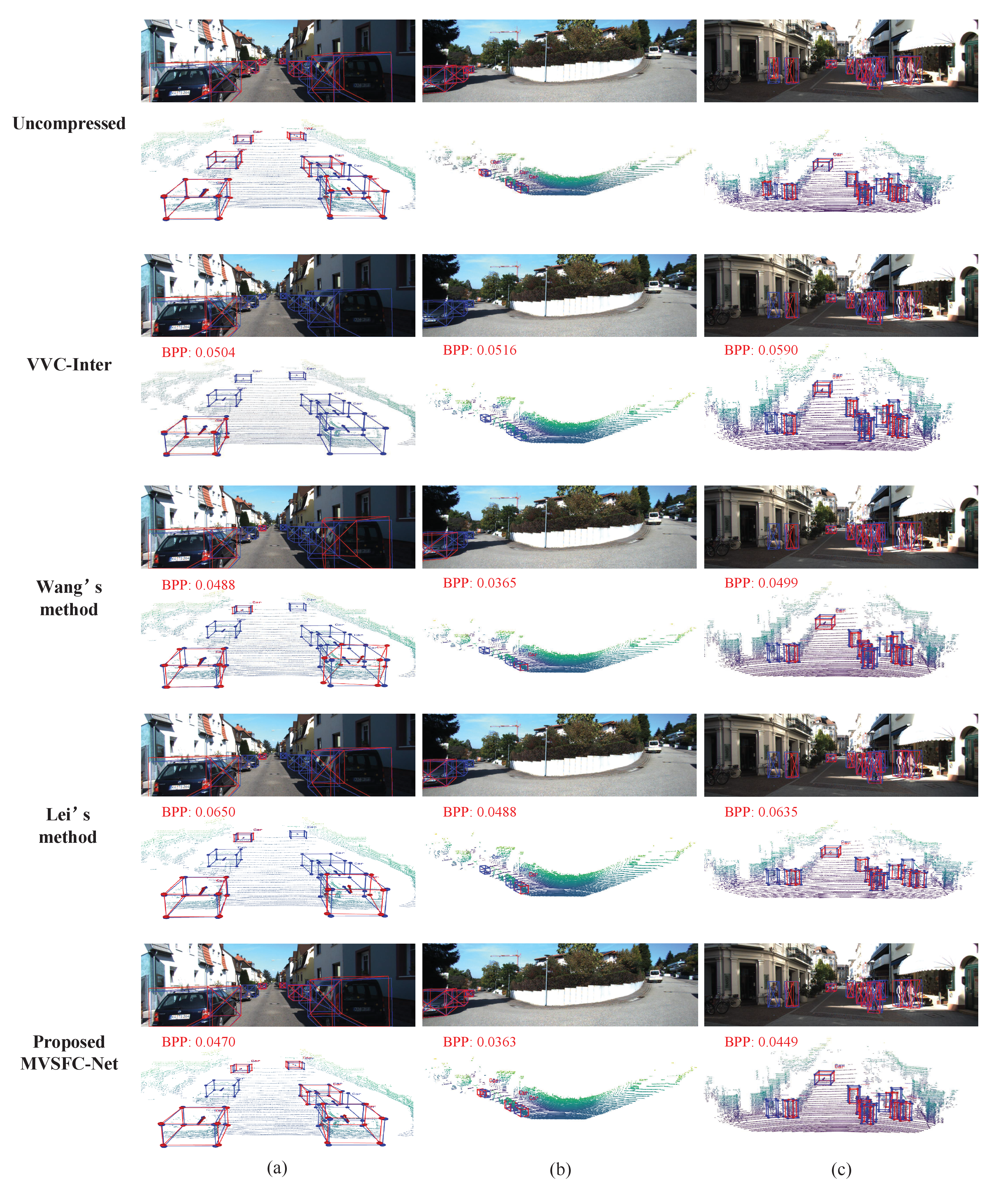}
	\caption{The visual comparison of 3D detection results in RGB images and 3D space. \textcolor[rgb]{0,0,0}{The blue bounding box and red bounding box denotes  the ground truth result and predicted result, respectively. (a) The 65-th left image in the validation set. (b) The 147-th left image in the validation set. (c) The 157-th left image in the validation set.}}
	
	\label{fig4}
\end{figure*}

\subsubsection{Visual Comparison}
To more intuitively prove the advantage of the proposed MVSFC-Net, the visual comparison of 3D detection results in RGB image and 3D space are presented in Fig. \ref{fig4}. In the figure, the ground truth 3D bounding boxes are drawn with the blue bounding boxes, while the red bounding boxes represent the 3D prediction results. As can be observed from the figure, compared with other comparison methods, the proposed MVSFC-Net is able to obtain more consistent 3D detection results with the uncompressed stereo images on both the Car category and the Pedestrian category, while having less bit-rate consumption.

\subsection{Ablation Study}
\begin{figure}[!t]
	\centering
	\includegraphics[width =8.0cm]{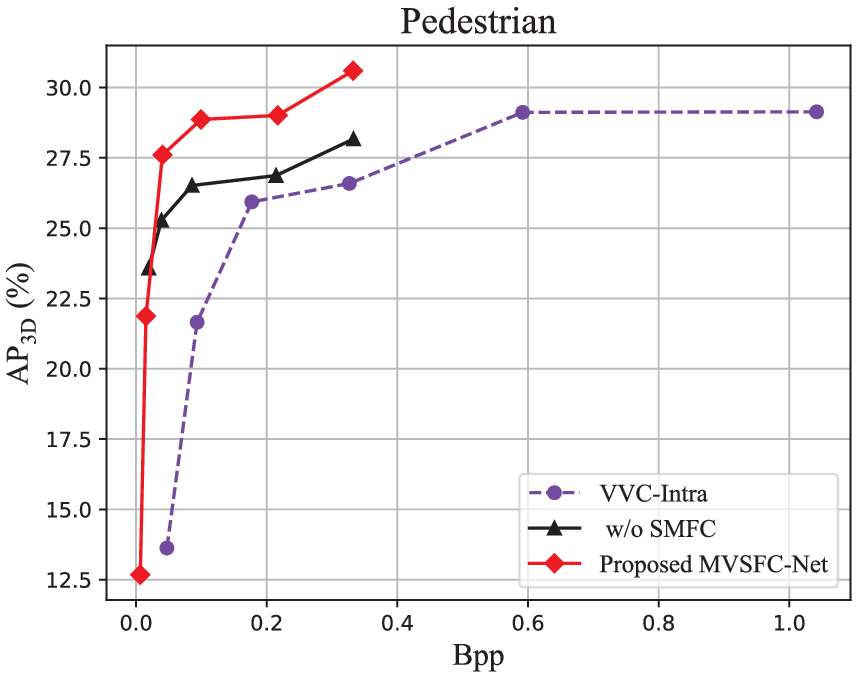}
	\caption{Ablation Results.}
	\label{fig5}
\end{figure}

The key of the proposed MVSFC-Net lies in efficient stereo features compression, as the stereo features exist enormously similar contents in images, viewpoints, and multiple scales. The SMFC module is designed to compress the stereo multi-scale features by effectively reducing the intra-, inter-view as well as cross-scale redundancies. To verify the effectiveness of the SMFC module, the SMFC module is removed in the MVSFC-Net, denoted as “w/o SMFC”. In the “w/o SMFC”, the stereo multi-scale features are separately transformed into the latent representations by down-sampling convolutions, and then the latent representations are concatenated together for entropy coding. As shown in Table \ref{table3}, when compared with the proposed MVSFC-Net, the network without the SMFC decreases the BD-rate reduction on average from 81.005\% to 70.401\%, and decreases the BD-$ \rm {AP_{3D}} $ from 9.368\% to 5.204\%. Meanwhile, as shown in Fig. \ref{fig5}, at the higher bitrate, the network without the SMFC drops more detection accuracy and is even close to the rate-distortion curve of VVC-Intra. This is because the inter-view and cross-scale redundancies in stereo multi-scale features are no longer effectively removed, resulting in a performance decrease.

\begin{table}[!t]
	\renewcommand{\arraystretch}{1.6}
	\footnotesize
	\centering
	\caption{BD-rate and BD-$ {\rm AP_{BEV}} $ Results of  MVSFC-Net Without SMFC Module}
	\setlength{\tabcolsep}{2.0mm}{
		\begin{tabular}
			{p{1.23cm}<{\centering}|c|c|c|c}
			\hline
			\hline
			\multirow {2}{*}{\textbf{Categories}} & \multicolumn{2}{c|}{\textbf{w/o SMFC}} &
			\multicolumn{2}{c}{\textbf{Proposed}}\\
			\cline{2-5}
			& \textbf{BD-rate} & \textbf{BD-$ \mathbf {AP_{3D}} $}  &  \textbf{BD-rate} & \textbf{BD-$ \mathbf {AP_{3D}} $}\\	
			
			\cline{1-5}
			\textbf{Car}
			& -71.535\% &	6.473\%  & \textbf{-78.254\%} & \textbf{12.477\%} \\ 
			\cline{1-5}
			\textbf{Pedestrian}
			&-69.266\% & 3.935\% & \textbf{-83.756\%}  &\textbf{6.259\%} \\
			\cline{1-5}
			\textbf{AVG}
			&-70.401\% & 5.204\% &	\textbf{-81.005\%}  & \textbf{9.368\%} \\
			\hline
			\hline
	\end{tabular}}%
	\label{table3}%
\end{table}%

\subsection{Computational Complexity Analysis}

We compare the computational complexity of the proposed MVSFC-Net with comparison algorithms in terms of number of floating-point operations (FLOPS), parameters, and encoding/decoding time. As shown in Table \ref {table4}, compared with the Lei's method \cite{2022-BCSIC} and Wang's method \cite{2021-CAS-Wang}, although the proposed MVSFC-Net has more parameters, the FLOPs of the proposed MVSFC-Net is reduced obviously, which is owing to the dealing with the feature-domain compact representation rather than the pixel-domain stereo images. As for the measurement of the encoding and decoding time, the QP in the traditional methods (\textit{i.e.}, VVC-Feature \cite{VVC-Feature}, VVC-Intra \cite{Image-anchor}, and VVC-Inter \cite{Image-anchor}) is set to 37, and the practical arithmetic coding is adopted in learning-based methods (\textit{i.e.}, the proposed MVSFC-Net, Lei's method \cite{2022-BCSIC}, and Wang's method \cite{2021-CAS-Wang}) for a fair comparison.  As shown in Table \ref {table4}, the proposed MVSFC-Net only consumes 0.396 s to encode the stereo images, and requires the lowest encoding time among all comparison methods. As for the decoding time, the proposed method  requires lowest decoding time when compared with learning-based methods, and has a higher decoding time than traditional methods due to the auto-regressive model in the entropy coding.

\begin{table}[!t]
	\renewcommand{\arraystretch}{1.6}
	\footnotesize
	\centering
	\caption{\textcolor[rgb]{0,0,0}{ Comparison Results of Computational Complexity}}
	\setlength{\tabcolsep}{0.5mm}{
		\begin{tabular}
			{p{2.4cm}<{\centering}|c|c|c|c}
			\hline
			\hline
			{\bfseries Method}&{\bfseries  FLOPs  (M)}&{\bfseries Params. (G) }&{\bfseries Enc-time (s)}&{\bfseries Dec-time (s) }\\
			\hline
			VVC-Feature\cite{VVC-Feature} & - & - & 365.716 & 0.080 \\
			\cline{1-5}
			VVC-Intra \cite{Image-anchor} & - & - & 46.324 & 0.051 \\
			\cline{1-5}
			VVC-Inter \cite{Image-anchor} & - & - & 37.480 &	0.048 \\
			\cline{1-5}
			\textcolor[rgb]{0,0,0}{Liu's method} \cite{2023-ICASSP-Liu} & 288.913 & 172.916 & 1.213 & 2.930 \\
			\cline{1-5}
			Wang's method \cite{2021-CAS-Wang} & 515.934 &	 124.188 & 0.601 & 4.864 \\
			\cline{1-5}
			Lei's method \cite{2022-BCSIC} & 	1172.725 & 136.722 & 1.283 &	5.384 \\
			\cline{1-5}			
			Proposed & 418.533 & 181.171 &	0.396 & 1.666 \\	
			\cline{1-5}	
			\hline
			\hline
	\end{tabular}}%
	\label{table4}%
\end{table}%

\section{Conclusion}\label{section5}
This paper firstly explores the potential framework for stereo 3D image coding for machine. In particular, a MVSFC-Net is proposed to extract, compress, and transmit compact visual features in a task-driven manner. In the MVSFC-Net, a stereo multi-scale feature compression module is proposed to remove spatial, inter-view, and cross-scale redundancies simultaneously by progressively generating the compact \textcolor[rgb]{0,0,0}{visual} joint representation. Experimental results prove that the MVSFC-Net presents more advanced compression capability in terms of the visual task performance, when compared with both state-of-the-art learning-based method and the latest traditional video coding standard VVC.


\ifCLASSOPTIONcaptionsoff
\newpage
\fi

\end{document}